\documentclass[sigconf]{acmart}

\usepackage{alltt}
\usepackage{listings}
\usepackage{xcolor}
\usepackage{amsmath, amssymb}
\usepackage{bm}
\usepackage{enumitem}

\lstdefinelanguage{python}{
  morestring=[b]',
  morestring=[b]""",
  morecomment=[l]\#,
  morekeywords={and,as,assert,break,class,continue,def,del,elif,else,except,False,finally,for,from,global,if,import,in,is,lambda,None,nonlocal,not,or,pass,raise,return,True,try,while,with,yield}
}

\newcommand*{\itsf}[1]{\textsf{\small\textit{#1}}}
\DeclareMathOperator*{\argmin}{argmin}
\DeclareMathOperator*{\matpar}{||}

\begin{document}

\settopmatter{printacmref=false} 
\renewcommand\footnotetextcopyrightpermission[1]{} 
\pagestyle{plain} 

\title{Deep Probabilistic Programming Languages: A Qualitative Study}

\author{Guillaume Baudart}
\affiliation{%
  \institution{IBM Research}
}
\email{guillaume.baudart@ibm.com}

\author{Martin Hirzel}
\affiliation{%
  \institution{IBM Research}
}
\email{hirzel@us.ibm.com}

\author{Louis Mandel}
\affiliation{%
  \institution{IBM Research}
}
\email{lmandel@us.ibm.com}


\begin{abstract}
Deep probabilistic programming languages try to combine the advantages
of deep learning with those of probabilistic programming languages. If
successful, this would be a big step forward in machine learning and
programming languages. Unfortunately, as of now, this new crop of
languages is hard to use and understand. This paper addresses this
problem directly by explaining deep probabilistic programming
languages and indirectly by characterizing their current strengths and
weaknesses.

\end{abstract}

%
%



\maketitle

\section{Introduction}\label{sec:intro}

A deep probabilistic programming language (PPL) is a language for
specifying both deep neural networks and probabilistic models.  In
other words, a deep PPL draws upon programming languages, Bayesian
statistics, and deep learning to ease the development of powerful
machine-learning applications.

For decades, scientists have developed probabilistic models in various
fields of exploration without the benefit of either dedicated programming
languages or deep neural networks~\cite{ghahramani_2015}. But since
these models involve Bayesian inference with often intractable
integrals, they sap the productivity of experts and are beyond the
reach of non-experts.  PPLs address this issue by letting users
express a probabilistic model as a program~\cite{gordon_et_al_2014}.
The program specifies how to generate output data by sampling latent
probability distributions.  The compiler checks this program for type
errors and translates it to a form suitable for an inference
procedure, which uses observed output data to fit the latent
distributions. Probabilistic models show great promise: they overtly
represent uncertainty~\cite{bornholt_mytkowicz_mckinley_2014} and
they have been demonstrated to enable explainable machine learning
even in the important but difficult case of small training
data~\cite{lake_salakhutdinov_tenenbaum_2015,rezende_et_al_2016,siddharth_et_al_2017}.

Over the last few years, machine learning with deep neural networks
(deep learning, DL) has become enormously popular. This is because in
several domains, DL solves what was previously a vexing
problem~\cite{domingos_2012}, namely manual feature
engineering. Each layer of a neural network can
be viewed as learning increasingly higher-level features. In other
words, the essence of DL is automatic hierarchical representation
learning~\cite{lecun_bengio_hinton_2015}.  Hence, DL powered recent
breakthrough results in accurate supervised large-data tasks such as
image recognition~\cite{krizhevsky_sutskever_hinton_2012} and natural
language translation~\cite{wu_et_al_2016}. Today, most DL is based on
frameworks that are well-supported, efficient, and expressive, such as
TensorFlow~\cite{abadi_et_al_2016} and PyTorch~\cite{pytorch}.
These frameworks provide automatic differentiation (users need not
manually calculate gradients for gradient descent), GPU support (to
efficiently execute vectorized computations), and Python-based
embedded domain-specific languages~\cite{hudak_1998}.

Deep PPLs, which have emerged just
recently~\cite{pyro,siddharth_et_al_2017,tran_et_al_2017,pymc3}, aim to combine
the benefits of PPLs and DL. Ideally, programs in deep PPLs would
overtly represent uncertainty, yield explainable models, and require
only a small amount of training data; be easy to write in a
well-designed programming language; and match the breakthrough
accuracy and fast training times of DL.  Realizing all of these
promises would yield tremendous advantages.  Unfortunately, this is
hard to achieve. Some of the strengths of PPLs and DL are seemingly at
odds, such as explainability vs.\ automated feature engineering, or
learning from small data vs.\ optimizing for large data. Furthermore,
the barrier to entry for work in deep PPLs is high, since it requires
non-trivial background in fields as diverse as statistics, programming
languages, and deep learning. To tackle this problem, this paper characterizes
deep PPLs, thus lowering the barrier to entry, providing a
programming-languages perspective early when it can make a difference,
and shining a light on gaps that the community should try to address.

This paper uses the Stan PPL as a representative of the state of the
art in regular (not deep) PPLs~\cite{carpenter_et_al_2017}. Stan is a
main-stream, mature, and widely-used PPL: it is maintained by
a large group of developers, has a yearly StanCon conference, and has an
active forum. Stan is Turing complete and has its own
stand-alone syntax and semantics, but provides bindings for several
languages including Python.

Most importantly, this paper uses Edward~\cite{tran_et_al_2017} and
Pyro~\cite{pyro} as representatives of the state of the art in deep
PPLs. Edward is based on TensorFlow and Pyro is based on PyTorch.
Edward was first released in mid-2016 and has a single main
maintainer, who is focusing on a new version.  Pyro is a much newer
framework (released late 2017), but seems to be very responsive to
community~questions.

This paper characterizes deep PPLs by explaining them
(Sections~\ref{sec:ppl_example}, \ref{sec:dl_example},
and~\ref{sec:both_example}), comparing them to each other and to
regular PPLs and DL frameworks (Section~\ref{sec:characterization}),
and envisioning next steps (Section~\ref{sec:conclusion}).
Additionally, the paper serves as a comparative
tutorial to both Edward and Pyro.  To this end, it presents
examples of increasing complexity written in both languages, using
deliberately uniform terminology and presentation style. By writing
this paper, we hope to help the research community contribute to the
exciting new field of deep PPLs, and ultimately, combine the strengths
of both DL and PPLs.

\section{Probabilistic Model Example}\label{sec:ppl_example}

This section explains PPLs using an example that is probabilistic but
not deep. The example, adapted from Section 9.1 of~\cite{barber_2012},
is about learning the bias of a coin. We picked this example because it
is simple, lets us introduce basic concepts, and shows how different
PPLs represent these concepts.

\begin{figure}
  \includegraphics[scale=0.4]{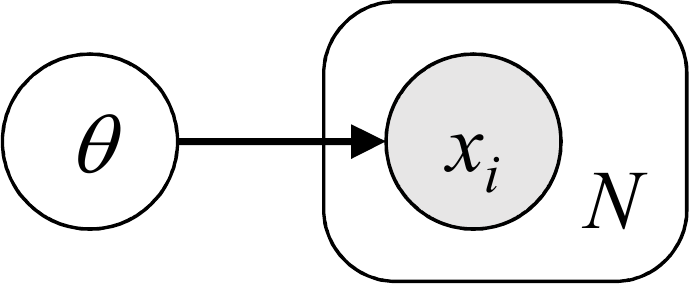}
  \caption{\label{fig:coin_graph}Graphical model for biased coin
    tosses. Circles represent random variables. The white circle
    for~$\theta$ indicates that it is latent and the gray circle
    for~$x_i$ indicates that it is observed.  The arrow represents
    dependency. The rounded rectangle is a plate, representing~$N$
    distributions that are IID.}
\end{figure}

We write $x_i=1$ if the result of the $i$\textsuperscript{th} coin
toss is head and $x_i=0$ if it is tail. We assume that individual coin
tosses are independent and identically distributed (IID) and that each
toss follows a \emph{Bernoulli} distribution with parameter $\theta$:
\mbox{$p(x_i=1\mid\theta)=\theta$} and
\mbox{$p(x_i=0\mid\theta)=1-\theta$}.  The \emph{latent} (i.e.,
unobserved) variable $\theta$ is the bias of the coin.  The task is to
infer $\theta$ given the results of previously observed coin tosses,
that is, \mbox{$p(\theta \mid x_1, x_2, \dots, x_N)$}.
Figure~\ref{fig:coin_graph} shows the corresponding \emph{graphical
  model}.  The model is \emph{generative}: once the distribution of
the latent variable has been inferred, one can draw samples to
generate data points similar to the observed data.

\bigskip

We now present this simple example in Stan, Edward, and Pyro. In all
these languages we follow a Bayesian approach: the programmer first
defines a probabilistic model of the problem.  Assumptions are encoded
with \emph{prior} distributions over the variables of the model. Then
the programmer launches an \emph{inference} procedure to automatically
compute the \emph{posterior} distributions of the parameters of the
model based on observed data. In other words, inference adjusts the
prior distribution using the observed data to give a more precise
model.  Compared to other machine-learning models such as deep neural
networks, the result of a probabilistic program is a probability
distribution, which allows the programmer to explicitly visualize and
manipulate the uncertainty associated with a result. This overt
uncertainty is an advantage of PPLs.  Furthermore, a probabilistic
model has the advantage that it directly describes the corresponding
world based on the programmer's knowledge. Such descriptive models are
more explainable than deep neural networks, whose representation is
big and does not overtly resemble the world they model.

\begin{figure*}
\hspace*{-10mm}\begin{tabular}{@{}ccc@{}}
\begin{minipage}[b]{0.3\textwidth}
\begin{lstlisting}[language=python]
# Model
coin_code = """
   data {
     int<lower=0,upper=1> x[10];
   }
   parameters {
     real<lower=0,upper=1> theta;
   }
   model {
     theta ~ uniform(0,1);
     for (i in 1:10)
        x[i] ~ bernoulli(theta);
   }"""
# Data
data = {'x': [0, 1, 0, 0, 0, 0, 0, 0, 0, 1]}
# Inference
fit = pystan.stan(model_code=coin_code,
                            data=data, iter=1000)
# Results
samples = fit.extract()['theta']
print("Posterior mean:", np.mean(samples))
print("Posterior stddev:", np.std(samples))
\end{lstlisting}
\end{minipage}
&
\begin{minipage}[b]{0.3\textwidth}
\begin{lstlisting}[language=python]
# Model
theta = Uniform(0.0, 1.0)
x = Bernoulli(probs=theta, sample_shape=10)
# Data
data = np.array([0, 1, 0, 0, 0, 0, 0, 0, 0, 1])
# Inference
qtheta = Empirical(
                   tf.Variable(tf.ones(1000) * 0.5))
inference = ed.HMC({theta: qtheta},
                                   data={x: data})
inference.run()
# Results
mean, stddev = ed.get_session().run(
                    [qtheta.mean(),qtheta.stddev()])
print("Posterior mean:", mean)
print("Posterior stddev:", stddev)
\end{lstlisting}
\end{minipage}
&
\begin{minipage}[b]{0.3\textwidth}
\begin{lstlisting}[language=python]
# Model
def coin():
   theta = pyro.sample("theta", Uniform(
                                      Variable(torch.Tensor([0])),
                                      Variable(torch.Tensor([1])))
   pyro.sample("x", Bernoulli(
                                  theta * Variable(torch.ones(10)))
# Data
data = {"x": Variable(torch.Tensor(
                                       [0, 1, 0, 0, 0, 0, 0, 0, 0, 1]))}
# Inference
cond = pyro.condition(coin, data=data)
sampler = pyro.infer.Importance(cond,
                                                      num_samples=1000)
post = pyro.infer.Marginal(sampler, sites=["theta"])
# Result
samples = [post()["theta"].data[0] for _ in range(1000)]
print("Posterior mean:", np.mean(samples))
print("Posterior stddev:", np.std(samples))
\end{lstlisting}
\end{minipage}\\
(a) Stan & (b) Edward & (c) Pyro
\end{tabular}
\vspace*{-2mm}
\caption{\label{fig:coin} Probabilistic model for learning the bias of a coin.}
\end{figure*}

Figure~\ref{fig:coin}(a) solves the biased-coin task in
Stan, a well-established PPL~\cite{carpenter_et_al_2017}. This example
uses PyStan, the Python interface for Stan. \mbox{Lines 2-13} contain
code in Stan syntax in a multi-line Python string.  The \itsf{data}
block introduces \emph{observed} random variables, which are
placeholders for concrete data to be provided as input to the
inference procedure, whereas the \itsf{parameters} block introduces
\emph{latent} random variables, which will not be provided and must be
inferred. Line~4 declares $x$ as a vector of ten \emph{discrete}
random variables, constrained to only take on values from a finite
set, in this case, either 0 or~1.  Line~7 declares $\theta$ as a
\emph{continuous} random variable, which can take on values from an
infinite set, in this case, real numbers between 0 and~1.  Stan uses
the tilde operator (\lstinline{~}) for sampling. Line~10 samples
$\theta$ from a \emph{uniform} distribution (same probability for all
values) between 0 and~1. Since $\theta$ is a latent variable, this
distribution is merely a \emph{prior} belief, which inference will
replace by a \emph{posterior} distribution.  Line~12 samples the $x_i$
from a Bernoulli distribution with parameter~$\theta$. Since the $x_i$
are observed variables, this sampling is really a check for how well
the model corresponds to data provided at inference time. One can
think of sampling an observed variable like an assertion in
verification~\cite{gordon_et_al_2014}.  Line~15 specifies the data and
\mbox{Lines 17-18} run the inference using the model and the data.
By default, Stan uses a form of Monte-Carlo sampling for
inference~\cite{carpenter_et_al_2017}.  \mbox{Lines 20-22} extract and
print the mean and standard deviation of the posterior distribution
of~$\theta$.

Figure~\ref{fig:coin}(b) solves the same task in
Edward~\cite{tran_et_al_2017}. \mbox{Line 2} samples~$\theta$ from the
prior distribution, and \mbox{Line 3} samples a vector of random
variables from a Bernoulli distribution of parameter~$\theta$, one for
each coin toss.
Line~5 specifies the data. \mbox{Lines 7-8} define a placeholder that
will be used by the inference procedure to compute the posterior
distribution of~$\theta$. The shape and size of the placeholder
depends on the inference procedure. Here we use the Hamiltonian
Monte-Carlo inference \lstinline{HMC}, the posterior distribution is
thus computed based on a set of random samples and follows an
empirical distribution. The size of the placeholder corresponds to the
number of random samples computed during inference. \mbox{Lines 9-11}
launch the inference. The inference takes as parameter the
prior:posterior pair \lstinline!theta:qtheta! and links the data to
the variable~$x$.  \mbox{Lines 13-16} extract and print the mean and
standard deviation of the posterior distribution of~$\theta$.

Figure~\ref{fig:coin}(c) solves the same task in Pyro~\cite{pyro}.
\mbox{Lines 2-7} define the model as a function \lstinline{coin}.
\mbox{Lines 3-5} sample~$\theta$ from the prior distribution, and
\mbox{Lines 6-7} sample a vector of random variable~$x$ from a
Bernoulli distribution of parameter~$\theta$. Pyro stores random
variables in a dictionary keyed by the first argument of the function
\lstinline{pyro.sample}. \mbox{Lines 9-10} define the data as a
dictionary. \mbox{Line 12} conditions the model using the data by
matching the value of the data dictionary with the random variables
defined in the model. \mbox{Lines 13-15} apply inference to this
conditioned model, using importance sampling.  Compared to Stan and
Edward, we first define a conditioned model with the observed data
before running the inference instead of passing the data as an
argument to the inference.  The inference returns a probabilistic
model, \lstinline{post}, that can be sampled to extract the mean and
standard deviation of the posterior distribution of~$\theta$ in
\mbox{Lines 17-19}.

\bigskip

The deep PPLs Edward and Pyro are built on top of two popular deep
learning frameworks TensorFlow~\cite{abadi_et_al_2016} and
PyTorch~\cite{pytorch}. They benefit from efficient computations over
large datasets, automatic differentiation, and optimizers provided by
these frameworks that can be used to efficiently implement inference
procedures.  As we will see in the next sections, this design choice
also reduces the gap between DL and probabilistic models, allowing the
programmer to combine the two.  On the other hand, this choice leads
to piling up abstractions (Edward/TensorFlow/Numpy/Python or
Pyro/PyTorch/Numpy/Python) that can complicate the code. We defer a
discussion of these towers of abstraction to
Section~\ref{sec:characterization}.

\begin{figure*}
\begin{tabular}{@{}cc@{}}
\begin{minipage}[b]{0.4\textwidth}
\begin{lstlisting}[language=python]
# Inference Guide
qalpha = tf.Variable(1.0)
qbeta = tf.Variable(1.0)
qtheta = Beta(qalpha, qbeta)
# Inference
inference = ed.KLqp({theta: qtheta}, {x: data})
inference.run()
\end{lstlisting}
\end{minipage}
&
\begin{minipage}[b]{0.6\textwidth}
\begin{lstlisting}[language=python]
# Inference Guide
def guide():
    qalpha = pyro.param("qalpha", Variable(torch.Tensor([1.0]), requires_grad=True))
    qbeta = pyro.param("qbeta", Variable(torch.Tensor([1.0]), requires_grad=True))
    pyro.sample("theta", Beta(qalpha, qbeta))
# Inference    
svi = SVI(cond, guide, Adam({}), loss="ELBO", num_particles=7)
for step in range(1000):
    svi.step()
\end{lstlisting}
\end{minipage}\\
(a) Edward & (b) Pyro
\end{tabular}
\vspace*{-2mm}
\caption{\label{fig:coin_vi} Variational Inference for learning the bias of a coin.}
\end{figure*}

\subsection*{Variational Inference}

Inference, for Bayesian models, computes the posterior distribution of
the latent parameters $\theta$ given a set of observations~$\bm{x}$,
that is,~${p(\theta \mid \bm{x})}$. For complex models, computing the
exact posterior distribution can be costly or even
intractable. \emph{Variational inference} turns the inference problem
into an optimization problem and tends to be faster and more adapted
to large datasets than sampling-based Monte-Carlo
methods~\cite{blei2017variational}.

Variational infence tries to find the member~$q^*(\theta)$ of a
family~$\mathcal{Q}$ of simpler distribution over the latent variables
that is the closest to the true posterior $p(\theta \mid \bm{x})$. The
fitness of a candidate is measured using the Kullback-Leibler (KL)
divergence from the true posterior, a similarity measure between
probability distributions.

$$
q^*(\theta) = \argmin_{q(\theta) \in \mathcal{Q}} \mbox{KL}(q(\theta) \matpar p(\theta \mid \bm{x}))
$$

It is up to the programmer to choose a family of candidates, or
\emph{guides}, that is sufficiently expressive to capture a close
approximation of the true posterior, but simple enough to make the
optimization problem tractable.

Both Edward and Pyro support variational inference.
Figure~\ref{fig:coin_vi} shows how to adapt Figure~\ref{fig:coin} to
use it.  In Edward
(Figure~\ref{fig:coin_vi}(a)), the programmer defines the family of
guides by changing the shape of the placeholder used in the
inference. \mbox{Lines 2-4} use a beta distribution with unknown
parameters~$\alpha$ and~$\beta$ that will be computed during
inference. \mbox{Lines 6-7} do variational inference using the
Kullback-Leibler divergence. In Pyro (Figure~\ref{fig:coin_vi}(b)),
this is done by defining a \emph{guide} function. \mbox{Lines 2-5}
also define a beta distribution with parameters~$\alpha$
and~$\beta$. \mbox{Lines 7-9} do inference using Stochastic
Variational Inference, an optimized algorithm for variational
inference. Both Edward and Pyro rely on the underlying framework to
solve the optimization problem. Probabilistic inference thus closely
follows the scheme used for training procedures of DL models.

\bigskip

This section gave a high-level introduction to PPLs and introduced
basic concepts (generative models, sampling, prior and posterior,
latent and observed, discrete and continuous).
Next, we turn our attention to deep learning.

\section{Probabilistic Models in DL}\label{sec:dl_example}

\begin{figure}
  \includegraphics[width=\columnwidth]{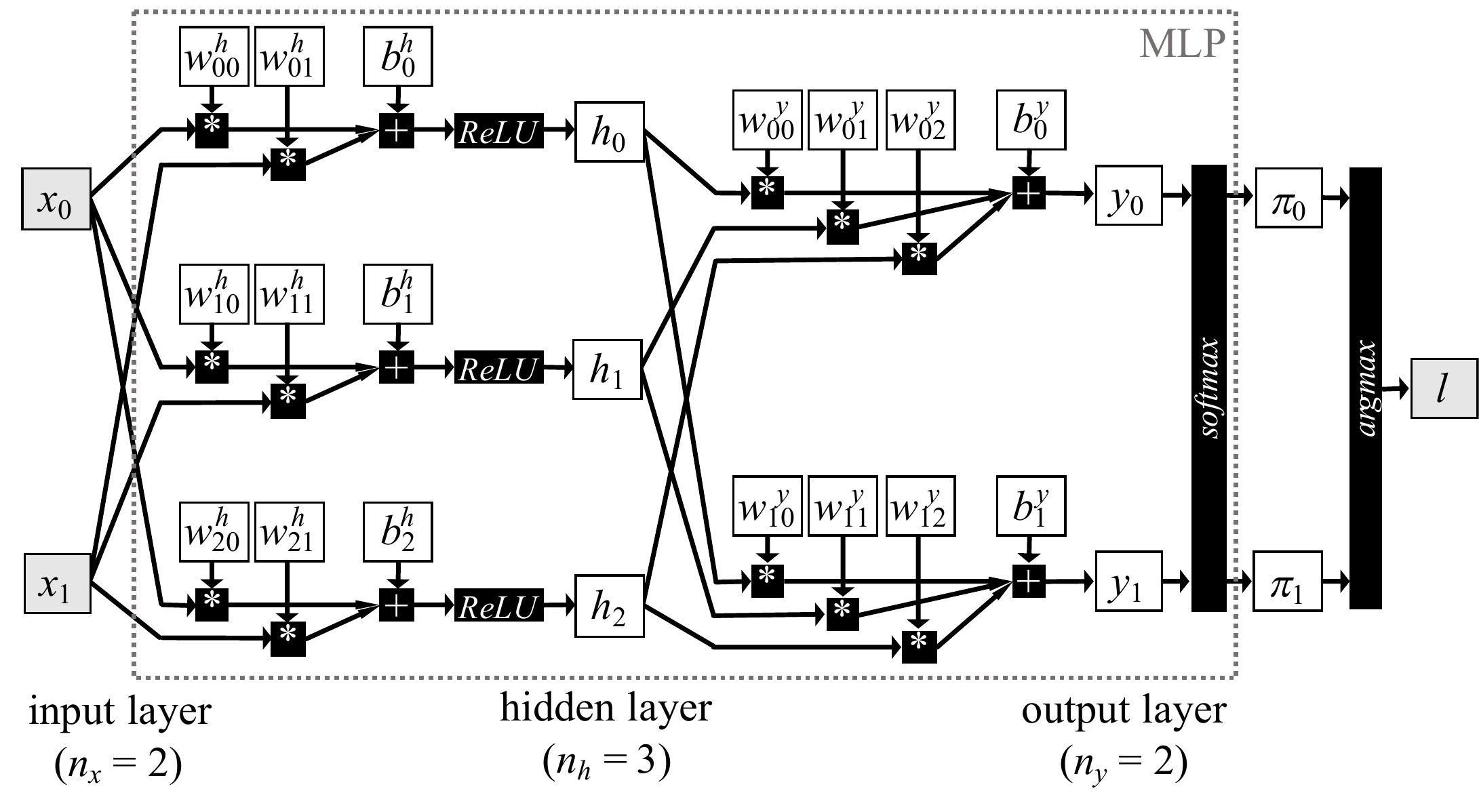}
  \centerline{(a) Computational graph for non-probabilistic MLP.}\\[1em]
  \includegraphics[scale=0.35]{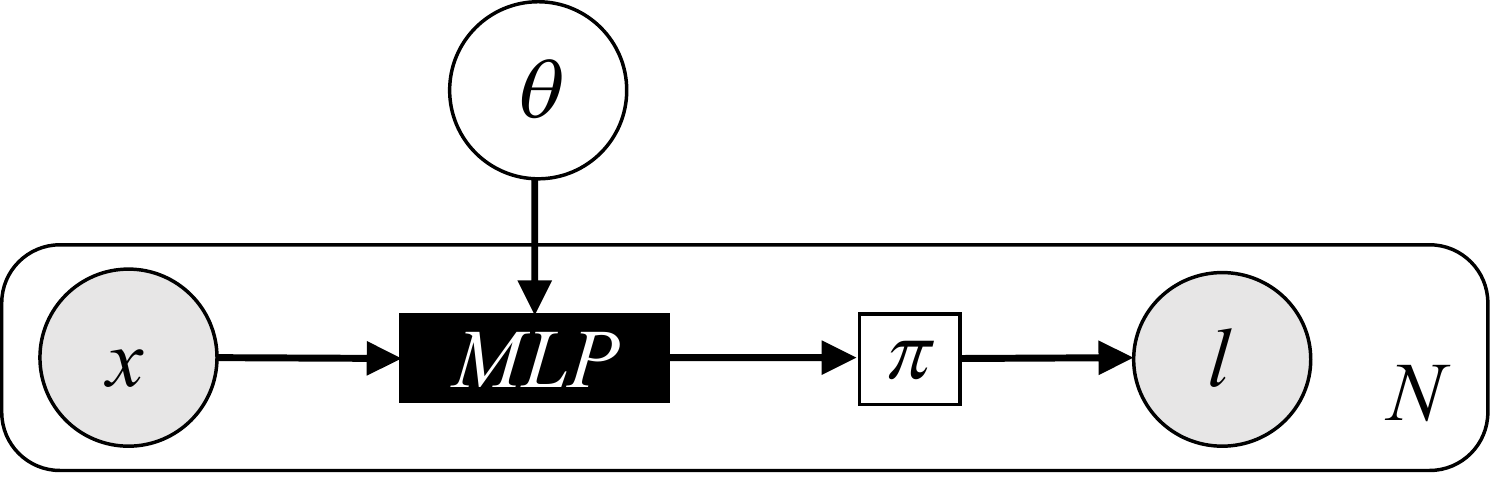}
  \centerline{(b) Graphical model for probabilistic MLP.}
  \caption{\label{fig:mlp_graph}Multi-layer perceptron (MLP) for
    classifying images. Circles and squares are probabilistic and
    non-probabilistic variables.  Black rectangles are pure
    functions. Arrows represent dependencies and forward data flow.}
  \vspace*{-3mm}
\end{figure}

\begin{figure*}
\hspace*{-6mm}\begin{tabular}{@{}c@{\hspace*{5mm}}c@{}}
\begin{minipage}[b]{0.45\textwidth}
\begin{lstlisting}[language=python]
batch_size, nx, nh, ny = 128, 28 * 28, 1024, 10
# Model
x = tf.placeholder(tf.float32, [batch_size, nx])
l = tf.placeholder(tf.int32, [batch_size])
def mlp(theta, x):
    h = tf.nn.relu(tf.matmul(x, theta["Wh"]) + theta["bh"])
    yhat = tf.matmul(h, theta["Wy"]) + theta["by"]
    log_pi = tf.nn.log_softmax(yhat)
    return log_pi
theta = {
    'Wh': Normal(loc=tf.zeros([nx, nh]), scale=tf.ones([nx, nh])),
    'bh': Normal(loc=tf.zeros(nh), scale=tf.ones(nh)),
    'Wy': Normal(loc=tf.zeros([nh, ny]), scale=tf.ones([nh, ny])),
    'by': Normal(loc=tf.zeros(ny), scale=tf.ones(ny)) }
lhat = Categorical(logits=mlp(theta, x))
# Inference Guide
def vr(*shape):
    return tf.Variable(tf.random_normal(shape))
qtheta = {
    'Wh': Normal(loc=vr(nx, nh), scale=tf.nn.softplus(vr(nx, nh))),
    'bh': Normal(loc=vr(nh), scale=tf.nn.softplus(vr(nh))),
    'Wy': Normal(loc=vr(nh, ny), scale=tf.nn.softplus(vr(nh, ny))),
    'by': Normal(loc=vr(ny), scale=tf.nn.softplus(vr(ny))) }
# Inference
inference = ed.KLqp({ theta["Wh"]: qtheta["Wh"],
                                    theta["bh"]: qtheta["bh"],
                                    theta["Wy"]: qtheta["Wy"],
                                    theta["by"]: qtheta["by"] },
                                  data={lhat: l})
\end{lstlisting}
\end{minipage}
&
\begin{minipage}[b]{0.45\textwidth}
\begin{lstlisting}[language=python]
# Model
class MLP(nn.Module):
    def __init__(self):
        super(MLP, self).__init__()
        self.lh = torch.nn.Linear(nx, nh)
        self.ly = torch.nn.Linear(nh, ny)
    def forward(self, x):
        h = F.relu(self.lh(x.view((-1, nx))))
        log_pi = F.log_softmax(self.ly(h), dim=-1)
        return log_pi
mlp = MLP()
def v0s(*shape): return Variable(torch.zeros(*shape))
def v1s(*shape): return Variable(torch.ones(*shape))
def model(x, l):
    theta = { 'lh.weight': Normal(v0s(nh, nx), v1s(nh, nx)),
                   'lh.bias': Normal(v0s(nh), v1s(nh)),
                   'ly.weight': Normal(v0s(ny, nh), v1s(ny, nh)),
                   'ly.bias': Normal(v0s(ny), v1s(ny)) }
    lifted_mlp = pyro.random_module("mlp", mlp, theta)()
    pyro.observe("obs", Categorical(logits=lifted_mlp(x)), one_hot(l))
# Inference Guide
def vr(name, *shape):
    return pyro.param(name,
               Variable(torch.randn(*shape), requires_grad=True))
def guide(x, l):
    qtheta = {
       'lh.weight': Normal(vr("Wh_m", nh, nx), F.softplus(vr("Wh_s", nh, nx))),
       'lh.bias': Normal(vr("bh_m", nh), F.softplus(vr("bh_s", nh))),
       'ly.weight': Normal(vr("Wy_m", ny, nh), F.softplus(vr("Wy_s", ny, nh))),
       'ly.bias': Normal(vr("by_m", ny), F.softplus(vr("by_s", ny))) }
    return pyro.random_module("mlp", mlp, qtheta)()
# Inference
inference = SVI(model, guide, Adam({"lr": 0.01}), loss="ELBO")
\end{lstlisting}
\end{minipage}\\
(a) Edward & (b) Pyro
\end{tabular}
\vspace*{-2mm}
\caption{\label{fig:mlp_code} Probabilistic multilayer perceptron for
  classifying images.}
\end{figure*}

This section explains DL using an example of a deep neural network
and shows how to make that probabilistic. The task is multiclass
classification: given input features~$x$, e.g., an image of a
handwritten digit~\cite{lecun_cortes_burges_1998} comprising
\mbox{$n_x=28\cdot28$} pixels, predict a label~$l$,
e.g., one of \mbox{$n_y=10$} digits. Before we explain
how to solve this task using DL, let us clarify some terminology.  In
cases where DL uses different terminology from PPLs, this paper favors
the PPL terminology. So we say \emph{inference} for what DL calls
training; \emph{predicting} for what DL calls inferencing; and
\emph{observed variables} for what DL calls \emph{training data}. For
instance, the observed variables for inference in the
classifier task are the image~$x$ and label~$l$.

Among the many neural network architectures suitable for this task, we
chose a simple one: a multi-layer perceptron
(MLP~\cite{rumelhart_hinton_williams_1986}).  We start with the
non-probabilistic version. Figure~\ref{fig:mlp_graph}(a) shows an MLP
with a 2-feature input layer, a 3-feature hidden layer, and a
2-feature output layer; of course, this generalizes to wider (more
features) and deeper (more layers) MLPs. From left to right, there is
a fully-connected subnet where each input feature~$x_i$ contributes to
each hidden feature~$h_j$, multiplied with a weight $w_{ji}$ and
offset with a bias~$b_j$. The weights and biases are latent
variables. Treating the input, biases, and weights as vectors and a
matrix lets us cast this subnet in linear algebra \mbox{$xW^h+b^h$},
which can run efficiently via vector instructions on CPUs or GPUs.
Next, a rectified linear unit \mbox{$\mathit{ReLU}(z)=\max(0,z)$}
computes the hidden feature vector~$h$. The ReLU lets the MLP
discriminate input spaces that are not linearly separable. The hidden
layer exhibits both the advantage and disadvantage of deep learning:
automatically learned features that need not be hand-engineered but
would require non-trivial reverse engineering to explain in real-world
terms.  Next, another fully-connected subnet \mbox{$hW^y+b^y$}
computes the output layer~$y$. Then, the softmax computes a vector~$\pi$
of scores that add up to one. The higher the value of~$\pi_l$, the
more strongly the classifier predicts label~$l$. Using the output of
the MLP, the argmax extracts the label~$l$ with highest score.


Traditional methods to train such a neural network incrementally
update the latent parameters of the network to optimize a loss function via
gradient descent~\cite{rumelhart_hinton_williams_1986}.
In the case of hand-written digits, the loss function
is a distance metrics between observed and predicted labels. The
variables and computations are non-probabilistic, in that they use
concrete values rather than probability distributions.


\bigskip

Deep PPLs can express Bayesian neural networks with probabilistic
weights and biases~\cite{blundell_et_al_2015}. One way to visualize
this is by replacing rectangles with circles for latent variables in
Figure~\ref{fig:mlp_graph}(a) to indicate that they hold probability
distributions. Figure~\ref{fig:mlp_graph}(b) shows the corresponding
graphical model, where the latent variable~$\theta$ denotes all the
parameters of the MLP: $W^h$, $b^h$, $W^y$, $b^y$.

Bayesian inference starts from prior beliefs about the parameters
and learns distributions that fit observed data (such as, images and
labels). We can then sample concrete weights and biases to obtain a
concrete MLP. In fact, we do not need to stop at a single MLP: we can
sample an ensemble of as many MLPs as we like.  Then, we can feed a
concrete image to all the sampled MLPs to get their predictions,
followed by a vote.

Figure~\ref{fig:mlp_code}(a) shows the probabilistic MLP example in
Edward. \mbox{Lines 3-4} are placeholders for observed variables
(i.e., batches of images~$x$ and labels~$l$). \mbox{Lines 5-9} defines
the MLP parameterized by~$\theta$, a dictionary containing all the
network parameters.  \mbox{Lines 10-14} sample the parameters from the
prior distributions.  \mbox{Line 15} defines the output of the
network: a categorical distribution over all possible label values
parameterized by the output of the MLP. \mbox{Line 17-23} define the
guides for the latent variables, initialized with random
numbers. Later, the variational inference will update these during
optimization, so they will ultimately hold an approximation of the
posterior distribution after inference.  \mbox{Lines 25-29} set up the
inference with one prior:posterior pair for each parameter of the
network and link the output of the network to the observed data.

Figure~\ref{fig:mlp_code}(b) shows the same example in
Pyro. \mbox{Lines 2-11} contain the basic neural network, where
\lstinline{torch.nn.Linear} wraps the low-level linear algebra.
\mbox{Lines 3-6} declare the structure of the net, that is, the type
and dimension of each layer. \mbox{Lines 7-10} combine the layers to
define the network.  It is possible to use equivalent high-level
TensorFlow APIs for this in Edward as well, but we refrained from
doing so to illustrate the transition of the parameters to random
variables. \mbox{Lines 14-20} define the model.  \mbox{Lines 15-18}
sample priors for the parameters, associating them with object
properties created by \lstinline{torch.nn.Linear} (i.e., the weight
and bias of each layer). \mbox{Line 19} lifts the MLP definition from
concrete to probabilistic. We thus obtain a MLP where all parameters
are treated as random variables. \mbox{Line 20} conditions the model
using a categorical distribution over all possible label
values. \mbox{Lines 26-31} define the guide for the latent variables,
initialized with random numbers, just like in the Edward
version. Line~33 sets up the inference.

\begin{figure*}
\vspace*{-2mm}
\hspace*{-6mm}\begin{tabular}{@{}c@{\hspace*{5mm}}c@{}}
\begin{minipage}[b]{0.555\textwidth}
\begin{lstlisting}[language=python]
def predict(x):
  theta_samples = [ { "Wh": qtheta["Wh"].sample(), "bh": qtheta["bh"].sample(),
                                   "Wy": qtheta["Wy"].sample(), "by": qtheta["by"].sample() }
                                   for _ in range(args.num_samples) ]
  yhats = [ mlp(theta_samp, x).eval()
                  for theta_samp in theta_samples ]
  mean = np.mean(yhats, axis=0)
  return np.argmax(mean, axis=1)
\end{lstlisting}
\end{minipage}
&
\begin{minipage}[b]{0.35\textwidth}
\begin{lstlisting}[language=python]
def predict(x):
  sampled_models = [ guide(None)
                                    for _ in range (args.num_samples) ]
  yhats = [ model(Variable(x)).data
                  for model in sampled_models ]
  mean = torch.mean(torch.stack(yhats), 0)
  return np.argmax(mean, axis=1)
\end{lstlisting}
\end{minipage}\\
(a) Edward & (b) Pyro
\end{tabular}
\vspace*{-2mm}
\caption{\label{fig:mlp_predict} Predictions by the probabilistic multilayer perceptron.}
\end{figure*}

After the inference, Figure~\ref{fig:mlp_predict} shows how to use
the posterior distribution of the MLP parameters to classify unknown
data.
In Edward~(Figure~\ref{fig:mlp_predict}(a)), \mbox{Lines 2-4} draw
several samples of the parameters from the posterior distribution.
Then, \mbox{Lines 5-6} execute the MLP with each concrete model.
\mbox{Line 7} computes the score of a label as the average of
the scores returned by the MLPs. Finally, \mbox{Line 8} predicts
the label with the highest score.
In Pyro~(Figure~\ref{fig:mlp_predict}(b)), the prediction is done
similarly but we obtain multiple versions of the MLP by sampling the guide
(\mbox{Line 2-3}), not the parameters.

This section showed how to use probabilistic variables as building
blocks for a DL model.
Compared to non-probabilistic DL, this approach has the advantage of
reduced overfitting and accurately quantified
uncertainty~\cite{blundell_et_al_2015}. On the other hand, this
approach requires inference techniques, like variational inference,
that are more advanced than classic back-propagation.
The next section will present the dual approach, showing how to use
neural networks as building blocks for a probabilistic model.

\section{DL in Probabilistic Models}\label{sec:both_example}

\begin{figure*}
\hspace*{-6mm}\begin{tabular}{@{}c@{\hspace*{10mm}}c@{}}
\begin{minipage}[b]{0.45\textwidth}
  \centerline{\includegraphics[scale=0.4]{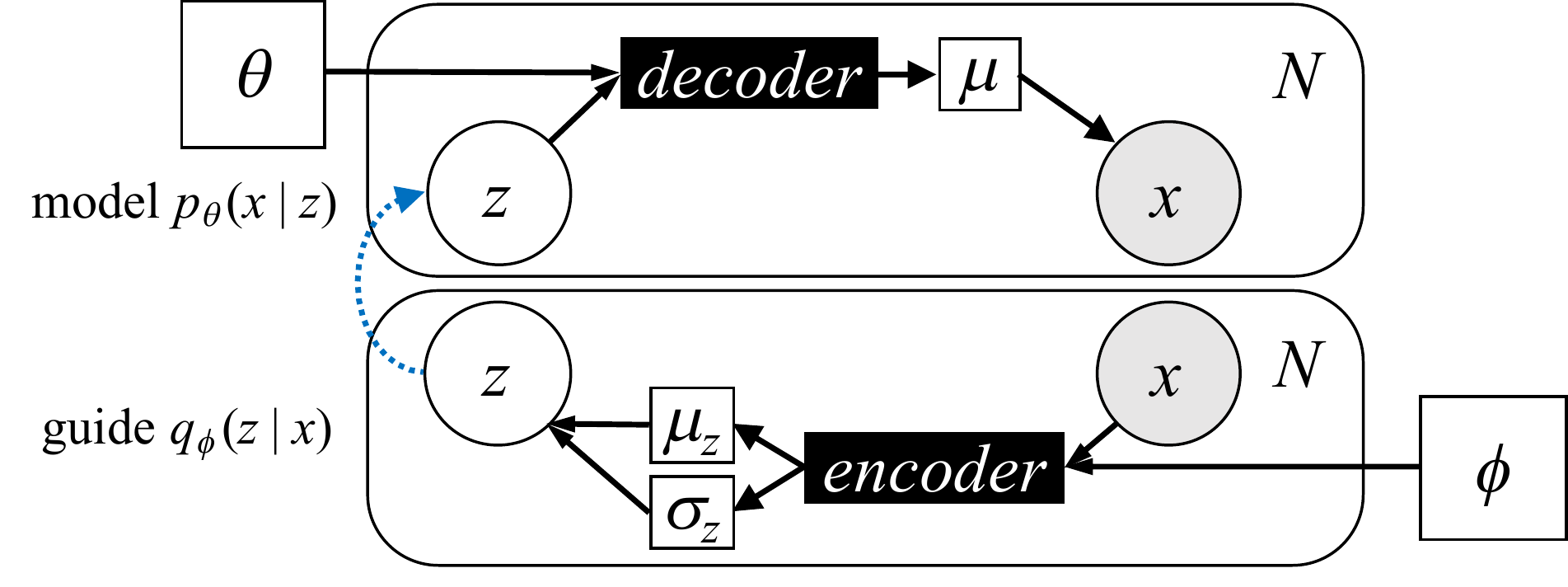}}
  \vspace*{2mm}
  \centerline{(a) Graphical models.}

  \vspace*{5mm}
\begin{lstlisting}[language=python]
batch_size, nx, nh, nz = 256, 28 * 28, 1024, 4
def vr(*shape): return tf.Variable(0.01 * tf.random_normal(shape))
# Model
X = tf.placeholder(tf.int32, [batch_size, nx])
def decoder(theta, z):
    hidden = tf.nn.relu(tf.matmul(z, theta['Wh']) + theta['bh'])
    mu = tf.matmul(hidden, theta['Wy']) + theta['by']
    return mu
theta = { 'Wh': vr(nz, nh),
               'bh': vr(nh),
               'Wy': vr(nh, nx),
               'by': vr(nx) }
z = Normal(loc=tf.zeros([batch_size, nz]), scale=tf.ones([batch_size, nz]))
logits = decoder(theta, z)
x = Bernoulli(logits=logits)
# Inference Guide
def encoder(phi, x):
    x = tf.cast(x, tf.float32)
    hidden = tf.nn.relu(tf.matmul(x, phi['Wh']) + phi['bh'])
    z_mu = tf.matmul(hidden, phi['Wy_mu']) + phi['by_mu']
    z_sigma = tf.nn.softplus(
        tf.matmul(hidden, phi['Wy_sigma']) + phi['by_sigma'])
    return z_mu, z_sigma
phi = { 'Wh': vr(nx, nh),
             'bh': vr(nh),
             'Wy_mu': vr(nh, nz),
             'by_mu': vr(nz),
             'Wy_sigma': vr(nh, nz),
             'by_sigma': vr(nz) }
loc, scale = encoder(phi, X)
qz = Normal(loc=loc, scale=scale)
# Inference
inference = ed.KLqp({z: qz}, data={x: X})
\end{lstlisting}
\end{minipage}
&
\begin{minipage}[b]{0.45\textwidth}
\begin{lstlisting}[language=python]
# Model
class Decoder(nn.Module):
    def __init__(self):
        super(Decoder, self).__init__()
        self.lh = nn.Linear(nz, nh)
        self.lx = nn.Linear(nh, nx)
        self.relu = nn.ReLU()
    def forward(self, z):
        hidden = self.relu(self.lh(z))
        mu = self.lx(hidden)
        return mu
decoder = Decoder()
def model(x):
    z_mu = Variable(torch.zeros(x.size(0), nz))
    z_sigma = Variable(torch.ones(x.size(0), nz))
    z = pyro.sample("z", dist.Normal(z_mu, z_sigma))
    pyro.module("decoder", decoder)
    mu = decoder.forward(z)
    pyro.sample("xhat", dist.Bernoulli(mu), obs=x.view(-1, nx))
# Inference Guide
class Encoder(nn.Module):
    def __init__(self):
        super(Encoder, self).__init__()
        self.lh = torch.nn.Linear(nx, nh)
        self.lz_mu = torch.nn.Linear(nh, nz)
        self.lz_sigma = torch.nn.Linear(nh, nz)
        self.softplus = nn.Softplus()
    def forward(self, x):
        hidden = F.relu(self.lh(x.view((-1, nx))))
        z_mu = self.lz_mu(hidden)
        z_sigma = self.softplus(self.lz_sigma(hidden))
        return z_mu, z_sigma
encoder = Encoder()
def guide(x):
    pyro.module("encoder", encoder)
    z_mu, z_sigma = encoder.forward(x)
    pyro.sample("z", dist.Normal(z_mu, z_sigma))
# Inference
inference = SVI(model, guide, Adam({"lr": 0.01}), loss="ELBO")

\end{lstlisting}
\end{minipage}\\
(b) Edward & (c) Pyro
\end{tabular}
\vspace*{-2mm}
\caption{\label{fig:vae}Variational autoencoder for encoding and decoding images.}
\end{figure*}

This section explains how deep PPLs can use non-probabilistic deep
neural networks as components in probabilistic models. The example
task is learning a vector-space representation. Such a representation
reduces the number of input dimensions to make other machine-learning
tasks more manageable by counter-acting the curse of
dimensionality~\cite{domingos_2012}. The observed random variable
is~$x$, for instance, an image of a hand-written digit with
\mbox{$n_x=28\cdot28$} pixels. The latent random variable is~$z$, the
vector-space representation, for instance, with \mbox{$n_z=4$}
features. Learning a vector-space representation is an unsupervised
problem, requiring only images but no labels. While not too useful on
its own, such a representation is an essential building block. For
instance, it can help in other image generation tasks, e.g., to
generate an image for a given writing
style~\cite{siddharth_et_al_2017}.  Furthermore, it can help learning
with small data, e.g., via a K-nearest neighbors approach in vector
space~\cite{babkin_et_al_2017}.

Each image~$x$ depends on the latent representation~$z$ in a
complex non-linear way (i.e., via a deep neural network). The task is
to learn this dependency between~$x$ and~$z$.  The top half of
Figure~\ref{fig:vae}(a) shows the corresponding graphical model. The
output of the neural network, named \emph{decoder}, is a vector~$\mu$
that parameterizes a Bernoulli distribution over each pixel in the
image~$x$. Each pixel is thus associated to a probability of being
present in the image.  Similarly to Figure~\ref{fig:mlp_graph}(b) the
parameter~$\theta$ of the decoder is global (i.e., shared by all data
points) and is thus drawn outside the plate. Compared to
Section~\ref{sec:dl_example} the network here is not probabilistic,
hence the square around~$\theta$.

The main idea of the
VAE~\cite{kingma_welling_2013,rezende_mohamed_wiestra_2014} is to use
variational inference to learn the latent representation. As for the
examples presented in the previous sections, we need to define a guide
for the inference. The guide maps each~$x$ to a latent
variable~$z$ via another neural network. The bottom half of
Figure~\ref{fig:vae}(a) shows the graphical model of the guide.  The
network, named \emph{encoder}, returns, for each image~$x$, the
parameters~$\mu_z$ and~$\sigma_z$ of a Gaussian distribution in the
latent space. Again the parameter~$\phi$ of the network is global and
not probabilistic. Then inference tries to learn good values for
parameter~$\theta$ and~$\phi$, simultaneously training the decoder and
the encoder, according to the data and the prior beliefs on the latent
variables (e.g., Gaussian distribution).

After the inference, we can generate a latent representation of an
image with the encoder and reconstruct the image with the decoder. The
similarity of the two images gives an indication of the success of the
inference.  The model and the guide together can thus be seen as an
autoencoder, hence the term \emph{variational autoencoder}.

\bigskip

Figure~\ref{fig:vae}(b) shows the VAE examples in Edward.  \mbox{Lines
  4-12} define the decoder: a simple 2-layers neural network similar
to the one presented in Section~\ref{sec:dl_example}. The parameter
$\theta$ is initialized with random noise.
\mbox{Line 13} samples the priors for the latent variable~$z$ from a
Gaussian distribution. \mbox{Lines 14-15} define the dependency
between~$x$ and~$z$, as a Bernoulli distribution parameterized by the
output of the decoder.
\mbox{Lines 17-29} define the encoder: a neural network with one
hidden layer and two distinct output layers for~$\mu_z$
and~$\sigma_z$. The parameter $\phi$ is also initialized with random
noise.
\mbox{Lines 30-31} define the inference guide for the latent
variable, that is, a Gaussian distribution parameterized by the outputs
of the encoder. \mbox{Line 33} sets up the inference matching the
prior:posterior pair for the latent variable and linking the data with
the output of the decoder.

Figure~\ref{fig:vae}(c) shows the VAE example in Pyro.  \mbox{Lines
  2-12} define the decoder.  \mbox{Lines 13-19} define the
model. \mbox{Lines 14-16} sample the priors for the latent
variable~$z$. \mbox{Lines 18-19} define the dependency between~$x$
and~$z$ via the decoder.
In contrast to Figure~\ref{fig:mlp_code}(b), the decoder is not
probabilistic, so there is no need for lifting the network.
\mbox{Lines 34-37} define the guide as in Edward linking~$z$ and~$x$
via the decoder defined \mbox{Lines 21-33}. \mbox{Line 39} sets up the
inference.

This example illustrates that we can embed non-probabilistic DL models
inside probabilistic programs and learn the parameters of the DL
models during inference.
Sections~\ref{sec:ppl_example}, \ref{sec:dl_example},
and~\ref{sec:both_example} were about explaining deep PPLs with
examples.  The next section is about comparing deep PPLs with each
other and with their potential.

\section{Characterization}\label{sec:characterization}

This section attempts to answer the following research question: At
this point in time, how well do deep PPLs live up to their potential?
Deep PPLs combine probabilistic models, deep learning, and programming
languages in an effort to combine their advantages. This section
explores those advantages grouped by pedigree and uses them to
characterize Edward and Pyro.

Before we dive in, some disclaimers are in order. First, both Edward and
Pyro are young, not mature projects with years of
improvements based on user experiences, and they enable new applications that
are still under active research. We should keep this in mind when we
criticize them.  On the positive side, early criticism can be a
positive influence. Second,
since getting even a limited number of example programs to support
direct side-by-side comparisons was non-trivial, we kept our study
qualitative. A respectable quantitative study would require more
programs and data sets. On the positive side, all of the code shown in
this paper actually runs. Third, doing an outside-in study risks
missing subtleties that the designers of Edward and Pyro may be more
expert in. On the positive side, the outside-in view resembles
what new users experience.

\subsection{Advantages from Probabilistic Models}

Probabilistic models support \emph{overt uncertainty}: they give not
just a prediction but also a meaningful probability. This is useful to
avoid uncertainty bugs~\cite{bornholt_mytkowicz_mckinley_2014}, track
compounding effects of uncertainty, and even make better exploration
decisions in reinforcement learning~\cite{blundell_et_al_2015}.  Both
Edward and Pyro support overt uncertainty well, see e.g.\ the lines
under the comment ``\lstinline{# Results}'' in Figure~\ref{fig:coin}.

Probabilistic models give users a \emph{choice of inference
  procedures}: the user has the flexibility to pick and configure
different approaches. Deep PPLs support two primary families of
inference procedures: those based on Monte-Carlo sampling and those
based on variational inference. Edward supports both and furthermore
flexible compositions, where different inference procedures are
applied to different parts of the model. Pyro supports primarily
variational inference and focuses less on Monte-Carlo
sampling. In comparison, Stan makes a form of Monte-Carlo sampling the
default, focusing on making it easy-to-tune in
practice~\cite{carpenter_2017}.

Probabilistic models can help with \emph{small data}: even when
inference uses only small amount of labeled data, there have been
high-profile cases where probabilistic models still make accurate
predictions~\cite{lake_salakhutdinov_tenenbaum_2015}. Working with
small data is useful to avoid the cost of hand-labeling, to improve
privacy, to build personalized models, and to do well on
underrepresented corners of a big-data task. The
intuition for how probabilistic models help is that they can make up
for lacking labeled data for a task by domain knowledge incorporated
in the model, by unlabeled data, or by labeled data for other tasks.
There are some promising initial successes of using deep probabilistic
programming on small
data \cite{rezende_et_al_2016,siddharth_et_al_2017}; at the same time,
this remains an active area of research.

Probabilistic models can support \emph{explainability}: when the
components of a probabilistic model correspond directly to concepts of
a real-world domain being modeled, predictions can include an
explanation in terms of those concepts. Explainability is useful when
predictions are consulted for high-stakes decisions, as well as for
transparency around bias~\cite{calmon_et_al_2017}.  Unfortunately, the
parameters of a deep neural network are just as opaque with as without
probabilistic programming. There is cause for hope though.
For instance, Siddharth et al.\ advocate disentangled representations
that help explainability~\cite{siddharth_et_al_2017}. Overall, the
jury is still out on the extent to which deep PPLs can leverage this
advantage from PPLs.

\subsection{Advantages from Deep Learning}

Deep learning is \emph{automatic hierarchical representation
  learning}~\cite{lecun_bengio_hinton_2015}: each unit in a deep
neural network can be viewed as an automatically learned feature.
Learning features automatically is useful to avoid the cost of
engineering features by hand. Fortunately, this DL advantage remains
true in the context of a deep PPL. In fact, a deep PPL makes the
trade-off between automated and hand-crafted features more flexible
than most other machine-learning approaches.

Deep learning can accomplish \emph{high accuracy}: for various tasks,
there have been high-profile cases where deep neural networks beat out
earlier approaches with years of research behind them. Arguably, the
victory of DL at the ImageNet competition in 2012 ushered in the
latest craze around DL~\cite{krizhevsky_sutskever_hinton_2012}.
Record-breaking accuracy is useful not just for publicity but also to
cross thresholds where practical deployments become desirable, for instance,
in machine translation~\cite{wu_et_al_2016}. Since a deep PPL can use
deep neural networks, in principle, it inherits this advantage from
DL~\cite{tran_et_al_2017}.  However, even non-probabilistic DL
requires tuning, and in our experience with the example programs in
this paper, the tuning burden is exacerbated with variational
inference.

Deep learning supports \emph{fast inference}: even for large models
and a large data set, the wall-clock time for a batch job to infer
posteriors is short. The fast-inference advantage is the result of the
back-propagation algorithm~\cite{rumelhart_hinton_williams_1986},
novel techniques for parallelization~\cite{niu_et_al_2011}
and data representation~\cite{gupta_et_al_2015}, and massive
investments in the efficiency of DL frameworks such as
TensorFlow and PyTorch, with
vectorization on CPU and GPU. Fast inference is useful for iterating
more quickly on ideas, trying more hyperparameter during, and
wasting fewer resources. Tran et al.\ measured the efficiency of the
Edward deep PPL, demonstrating that it does benefit from the
efficiency advantage of the underlying DL
framework~\cite{tran_et_al_2017}.

\subsection{Advantages from Programming Languages}

Programming language design is essential for \emph{composability}:
bigger models can be composed from smaller components. Composability
is useful for testing, team-work, and reuse. Conceptually, both
graphical probabilistic models and deep neural networks compose well.
On the other hand, some PPLs impose structure in a way that reduces
composability; fortunately, this can be
mitigated~\cite{gorinova_gordon_sutton_2018}. Both Edward and Pyro are
embedded in Python, and, as our example programs demonstrate, work
with Python functions and classes. For instance, users are not limited
to declaring all latent variables in one place; instead, they can
compose models, such as MLPs, with separately declared latent
variables. Edward and Pyro also work with higher-level DL framework
modules such as \lstinline{tf.layers.dense} or
\lstinline{torch.nn.Linear}, and Pyro even supports automatically
lifting those to make them probabilistic. Edward and Pyro also do not
prevent users from composing probabilistic models with
non-probabilistic code, but doing so requires care. For instance, when
Monte-Carlo sampling runs the same code multiple times, it is up to
the programmer to watch out for unwanted side-effects. One area where
more work is needed is the extensibility of Edward or Pyro
itself~\cite{carpenter_2017}. Finally, in addition to composing
models, Edward also emphasizes composing inference
procedures.

Not all PPLs have the same \emph{expressiveness}: some are Turing
complete, others not~\cite{gordon_et_al_2014}. For instance, BUGS is
not Turing complete, but has nevertheless been highly
successful~\cite{gilks_thomas_spiegelhalter_1994}.  The field of deep
probabilistic programming is too young to judge which levels of expressiveness
are how useful.  Edward and Pyro are both Turing complete. However,
Edward makes it harder to express while-loops and conditionals than
Pyro. Since Edward builds on TensorFlow, the user must use special
APIs to incorporate dynamic control constructs into the computational
graph. In contrast, since Pyro builds on PyTorch, it can use native
Python control constructs, one of the advantages of dynamic DL
frameworks~\cite{neubig_et_al_2017}.

Programming language design affects \emph{conciseness}: it determines
whether a model can be expressed in few lines of code. Conciseness is
useful to make models easier to write and, when used in good taste,
easier to read. In our code examples, Edward is more concise than
Pyro. Pyro arguably trades conciseness for structure, making heavier
use of Python classes and functions. Wrapping the model and guide into
functions allows compiling them into co-routines, an ingredient for
implementing efficient inference procedures~\cite{goodman_stuhlmuller_2014}.
In both Edward and Pyro, conciseness is hampered by the Bayesian
requirement for explicit priors and by the variational-inference
requirement for explicit guides.

Programming languages can offer \emph{watertight abstractions}: they
can abstract away lower-level concepts and prevent them from leaking
out, for instance, using types and constraints~\cite{carpenter_2017}.
Consider the expression \mbox{$xW^{[1]}+b^{[1]}$} from
Section~\ref{sec:dl_example}. At face value, this looks like eager
arithmetic on concrete scalars, running just once in the forward
direction. But actually, it may be lazy (building a computational
graph) arithmetic on probability distributions (not concrete values)
of tensors (not scalars), running several times (for different
Monte-Carlo samples or data batches), possibly in the backward
direction (for back-propagation of gradients). Abstractions are
useful to reduce the cognitive burden on the programmer, but only if
they are watertight. Unfortunately, abstractions in deep PPLs are
leaky. Our code examples directly invoke features from several layers
of the technology stack (Edward or Pyro, on TensorFlow or PyTorch, on NumPy, on
Python). Furthermore, we found that error messages rarely
refer to source code constructs. For instance, names of
Python variables are erased from the computational graph, making it
hard to debug tensor dimensions, a common cause for mistakes. It does
not have to be that way. For instance, DL frameworks are good at
hiding the abstraction of back-propagation.
More work is required to make deep PPL abstractions more watertight.

\section{Conclusion and Outlook}\label{sec:conclusion}

This paper is a study of two deep PPLs, Edward and Pyro. The study is
qualitative and driven by code examples. This paper explains how to
solve common tasks, contributing side-by-side comparisons of Edward
and Pyro. The potential of deep PPLs is to combine the advantages of
probabilistic models, deep learning, and programming languages.
In addition to comparing Edward and Pyro to each other, this paper
also compares them to that potential. A quantitative study is left to future
work. Based on our experience, we confirm that
Edward and Pyro combine three advantages out-of-the-box: the overt
uncertainty of probabilistic models; the hierarchical representation
learning of DL; and the composability of programming languages.

Following are possible next steps in deep PPL research.

\begin{itemize}[leftmargin=\parindent]
  \item Choice of inference procedures: Especially Pyro should support
    Monte-Carlo methods at the same level as variational inference.
  \item Small data: While possible in theory, this has yet to be
    demonstrated on Edward and Pyro, with interesting data sets.
  \item High accuracy: Edward and Pyro need to be improved to simplify
    the tuning required to improve model accuracy.
  \item Expressiveness: While Turing complete in theory, Edward should
    adopt recent dynamic TensorFlow features for usability.
  \item Conciseness: Both Edward and Pyro would benefit from reducing
    the repetitive idioms of priors and guides.
  \item Watertight abstractions: Both Edward and Pyro fall short on
    this goal, necessitating more careful language design.
  \item Explainability: This is inherently hard with deep PPLs,
    necessitating more machine-learning innovation.
\end{itemize}

\noindent
In summary, deep PPLs show great promises and remain an active field
with many research opportunities.

\bibliographystyle{ACM-Reference-Format}
\bibliography{bibfile}


\begin{thebibliography}{33}


\ifx \showCODEN    \undefined \def \showCODEN     #1{\unskip}     \fi
\ifx \showDOI      \undefined \def \showDOI       #1{#1}\fi
\ifx \showISBNx    \undefined \def \showISBNx     #1{\unskip}     \fi
\ifx \showISBNxiii \undefined \def \showISBNxiii  #1{\unskip}     \fi
\ifx \showISSN     \undefined \def \showISSN      #1{\unskip}     \fi
\ifx \showLCCN     \undefined \def \showLCCN      #1{\unskip}     \fi
\ifx \shownote     \undefined \def \shownote      #1{#1}          \fi
\ifx \showarticletitle \undefined \def \showarticletitle #1{#1}   \fi
\ifx \showURL      \undefined \def \showURL       {\relax}        \fi
\providecommand\bibfield[2]{#2}
\providecommand\bibinfo[2]{#2}
\providecommand\natexlab[1]{#1}
\providecommand\showeprint[2][]{arXiv:#2}

\bibitem[\protect\citeauthoryear{Abadi, Barham, Chen, Chen, Davis, Dean, Devin,
  Ghemawat, Irving, Isard, Kudlur, Levenberg, Monga, Moore, Murray, Steiner,
  Tucker, Vasudevan, Warden, Wicke, Yu, and Zheng}{Abadi et~al\mbox{.}}{2016}]%
        {abadi_et_al_2016}
\bibfield{author}{\bibinfo{person}{Mart{\'i}n Abadi}, \bibinfo{person}{Paul
  Barham}, \bibinfo{person}{Jianmin Chen}, \bibinfo{person}{Zhifeng Chen},
  \bibinfo{person}{Andy Davis}, \bibinfo{person}{Jeffrey Dean},
  \bibinfo{person}{Matthieu Devin}, \bibinfo{person}{Sanjay Ghemawat},
  \bibinfo{person}{Geoffrey Irving}, \bibinfo{person}{Michael Isard},
  \bibinfo{person}{Manjunath Kudlur}, \bibinfo{person}{Josh Levenberg},
  \bibinfo{person}{Rajat Monga}, \bibinfo{person}{Sherry Moore},
  \bibinfo{person}{Derek~G. Murray}, \bibinfo{person}{Benoit Steiner},
  \bibinfo{person}{Paul Tucker}, \bibinfo{person}{Vijay Vasudevan},
  \bibinfo{person}{Pete Warden}, \bibinfo{person}{Martin Wicke},
  \bibinfo{person}{Yuan Yu}, {and} \bibinfo{person}{Xiaoqiang Zheng}.}
  \bibinfo{year}{2016}\natexlab{}.
\newblock \showarticletitle{{TensorFlow}: A System for Large-Scale Machine
  Learning}. In \bibinfo{booktitle}{\emph{Operating Systems Design and
  Implementation (OSDI)}}. \bibinfo{pages}{265--283}.
\newblock
\urldef\tempurl%
\url{https://www.usenix.org/conference/osdi16/technical-sessions/presentation/abadi}
\showURL{%
\tempurl}


\bibitem[\protect\citeauthoryear{Babkin, Chowdhury, Gliozzo, Hirzel, and
  Shinnar}{Babkin et~al\mbox{.}}{2017}]%
        {babkin_et_al_2017}
\bibfield{author}{\bibinfo{person}{Petr Babkin}, \bibinfo{person}{Md.
  Faisal~Mahbub Chowdhury}, \bibinfo{person}{Alfio Gliozzo},
  \bibinfo{person}{Martin Hirzel}, {and} \bibinfo{person}{Avraham Shinnar}.}
  \bibinfo{year}{2017}\natexlab{}.
\newblock \showarticletitle{Bootstrapping Chatbots for Novel Domains}. In
  \bibinfo{booktitle}{\emph{Workshop at NIPS on Learning with Limited Labeled
  Data (LLD)}}.
\newblock
\urldef\tempurl%
\url{https://lld-workshop.github.io/papers/LLD_2017_paper_10.pdf}
\showURL{%
\tempurl}


\bibitem[\protect\citeauthoryear{Barber}{Barber}{2012}]%
        {barber_2012}
\bibfield{author}{\bibinfo{person}{David Barber}.}
  \bibinfo{year}{2012}\natexlab{}.
\newblock \bibinfo{booktitle}{\emph{Bayesian Reasoning and Machine Learning}}.
\newblock \bibinfo{publisher}{Cambridge University Press}.
\newblock
\urldef\tempurl%
\url{http://www.cs.ucl.ac.uk/staff/d.barber/brml/}
\showURL{%
\tempurl}


\bibitem[\protect\citeauthoryear{Blei, Kucukelbir, and McAuliffe}{Blei
  et~al\mbox{.}}{2017}]%
        {blei2017variational}
\bibfield{author}{\bibinfo{person}{David~M. Blei}, \bibinfo{person}{Alp
  Kucukelbir}, {and} \bibinfo{person}{Jon~D. McAuliffe}.}
  \bibinfo{year}{2017}\natexlab{}.
\newblock \showarticletitle{Variational inference: A review for statisticians}.
\newblock \bibinfo{journal}{\emph{J. Amer. Statist. Assoc.}}
  \bibinfo{volume}{112}, \bibinfo{number}{518} (\bibinfo{year}{2017}),
  \bibinfo{pages}{859--877}.
\newblock
\urldef\tempurl%
\url{https://arxiv.org/abs/1601.00670}
\showURL{%
\tempurl}


\bibitem[\protect\citeauthoryear{Blundell, Cornebise, Kavukcuoglu, and
  Wierstra}{Blundell et~al\mbox{.}}{2015}]%
        {blundell_et_al_2015}
\bibfield{author}{\bibinfo{person}{Charles Blundell}, \bibinfo{person}{Julien
  Cornebise}, \bibinfo{person}{Koray Kavukcuoglu}, {and} \bibinfo{person}{Daan
  Wierstra}.} \bibinfo{year}{2015}\natexlab{}.
\newblock \showarticletitle{Weight Uncertainty in Neural Network}. In
  \bibinfo{booktitle}{\emph{International Conference on Machine Learning
  (ICML)}}. \bibinfo{pages}{1613--1622}.
\newblock
\urldef\tempurl%
\url{http://proceedings.mlr.press/v37/blundell15.html}
\showURL{%
\tempurl}


\bibitem[\protect\citeauthoryear{Bornholt, Mytkowicz, and McKinley}{Bornholt
  et~al\mbox{.}}{2014}]%
        {bornholt_mytkowicz_mckinley_2014}
\bibfield{author}{\bibinfo{person}{James Bornholt}, \bibinfo{person}{Todd
  Mytkowicz}, {and} \bibinfo{person}{Kathryn~S. McKinley}.}
  \bibinfo{year}{2014}\natexlab{}.
\newblock \showarticletitle{Uncertain<T>: A First-order Type for Uncertain
  Data}. In \bibinfo{booktitle}{\emph{Conference on Architectural Support for
  Programming Languages and Operating Systems (ASPLOS)}}.
  \bibinfo{pages}{51--66}.
\newblock
\urldef\tempurl%
\url{https://doi.org/10.1145/2541940.2541958}
\showURL{%
\tempurl}


\bibitem[\protect\citeauthoryear{Calmon, Wei, Vinzamuri, Ramamurthy, and
  Varshney}{Calmon et~al\mbox{.}}{2017}]%
        {calmon_et_al_2017}
\bibfield{author}{\bibinfo{person}{Flavio~P. Calmon}, \bibinfo{person}{Dennis
  Wei}, \bibinfo{person}{Bhanukiran Vinzamuri},
  \bibinfo{person}{Karthikeyan~Natesan Ramamurthy}, {and}
  \bibinfo{person}{Kush~R. Varshney}.} \bibinfo{year}{2017}\natexlab{}.
\newblock \showarticletitle{Optimized Pre-Processing for Discrimination
  Prevention}. In \bibinfo{booktitle}{\emph{Neural Information Processing
  Systems (NIPS)}}. \bibinfo{pages}{3995--4004}.
\newblock
\urldef\tempurl%
\url{http://papers.nips.cc/paper/6988-optimized-pre-processing-for-discrimination-prevention.pdf}
\showURL{%
\tempurl}


\bibitem[\protect\citeauthoryear{Carpenter}{Carpenter}{2017}]%
        {carpenter_2017}
\bibfield{author}{\bibinfo{person}{Bob Carpenter}.}
  \bibinfo{year}{2017}\natexlab{}.
\newblock \bibinfo{title}{Hello, world! Stan, {PyMC3}, and {Edward}}.
\newblock   (\bibinfo{year}{2017}).
\newblock
\urldef\tempurl%
\url{http://andrewgelman.com/2017/05/31/compare-stan-pymc3-edward-hello-world/}
\showURL{%
\tempurl}
\newblock
\shownote{(Retrieved February 2018).}


\bibitem[\protect\citeauthoryear{Carpenter, Gelman, Hoffman, Lee, Goodrich,
  Betancourt, Brubaker, Guo, Li, and Riddell}{Carpenter et~al\mbox{.}}{2017}]%
        {carpenter_et_al_2017}
\bibfield{author}{\bibinfo{person}{Bob Carpenter}, \bibinfo{person}{Andrew
  Gelman}, \bibinfo{person}{Matt Hoffman}, \bibinfo{person}{Daniel Lee},
  \bibinfo{person}{Ben Goodrich}, \bibinfo{person}{Michael Betancourt},
  \bibinfo{person}{Michael~A. Brubaker}, \bibinfo{person}{Jiqiang Guo},
  \bibinfo{person}{Peter Li}, {and} \bibinfo{person}{Allen Riddell}.}
  \bibinfo{year}{2017}\natexlab{}.
\newblock \showarticletitle{Stan: A probabilistic programming language}.
\newblock \bibinfo{journal}{\emph{Journal of Statistical Software}}
  \bibinfo{volume}{76}, \bibinfo{number}{1} (\bibinfo{year}{2017}),
  \bibinfo{pages}{1--37}.
\newblock
\urldef\tempurl%
\url{https://www.jstatsoft.org/article/view/v076i01}
\showURL{%
\tempurl}


\bibitem[\protect\citeauthoryear{Domingos}{Domingos}{2012}]%
        {domingos_2012}
\bibfield{author}{\bibinfo{person}{Pedro Domingos}.}
  \bibinfo{year}{2012}\natexlab{}.
\newblock \showarticletitle{A Few Useful Things to Know About Machine
  Learning}.
\newblock \bibinfo{journal}{\emph{Communications of the ACM (CACM)}}
  \bibinfo{volume}{55}, \bibinfo{number}{10} (\bibinfo{date}{Oct.}
  \bibinfo{year}{2012}), \bibinfo{pages}{78--87}.
\newblock
\urldef\tempurl%
\url{https://doi.org/10.1145/2347736.2347755}
\showURL{%
\tempurl}


\bibitem[\protect\citeauthoryear{Facebook}{Facebook}{2016}]%
        {pytorch}
\bibfield{author}{\bibinfo{person}{Facebook}.} \bibinfo{year}{2016}\natexlab{}.
\newblock \bibinfo{title}{PyTorch}.
\newblock   (\bibinfo{year}{2016}).
\newblock
\urldef\tempurl%
\url{http://pytorch.org/}
\showURL{%
\tempurl}
\newblock
\shownote{(Retrieved February 2018).}


\bibitem[\protect\citeauthoryear{Ghahramani}{Ghahramani}{2015}]%
        {ghahramani_2015}
\bibfield{author}{\bibinfo{person}{Zoubin Ghahramani}.}
  \bibinfo{year}{2015}\natexlab{}.
\newblock \showarticletitle{Probabilistic machine learning and artificial
  intelligence}.
\newblock \bibinfo{journal}{\emph{Nature}} \bibinfo{volume}{521},
  \bibinfo{number}{7553} (\bibinfo{date}{May} \bibinfo{year}{2015}),
  \bibinfo{pages}{452--459}.
\newblock
\urldef\tempurl%
\url{https://www.nature.com/articles/nature14541}
\showURL{%
\tempurl}


\bibitem[\protect\citeauthoryear{Gilks, Thomas, and Spiegelhalter}{Gilks
  et~al\mbox{.}}{1994}]%
        {gilks_thomas_spiegelhalter_1994}
\bibfield{author}{\bibinfo{person}{W.~R. Gilks}, \bibinfo{person}{A. Thomas},
  {and} \bibinfo{person}{D.~J. Spiegelhalter}.}
  \bibinfo{year}{1994}\natexlab{}.
\newblock \showarticletitle{A Language and Program for Complex Bayesian
  Modelling}.
\newblock \bibinfo{journal}{\emph{The Statistician}} \bibinfo{volume}{43},
  \bibinfo{number}{1} (\bibinfo{date}{Jan.} \bibinfo{year}{1994}),
  \bibinfo{pages}{169--177}.
\newblock


\bibitem[\protect\citeauthoryear{Goodman and Stuhlm{\"u}ller}{Goodman and
  Stuhlm{\"u}ller}{2014}]%
        {goodman_stuhlmuller_2014}
\bibfield{author}{\bibinfo{person}{Noah~D. Goodman} {and}
  \bibinfo{person}{Andreas Stuhlm{\"u}ller}.} \bibinfo{year}{2014}\natexlab{}.
\newblock \bibinfo{title}{The Design and Implementation of Probabilistic
  Programming Languages}.
\newblock   (\bibinfo{year}{2014}).
\newblock
\urldef\tempurl%
\url{http://dippl.org}
\showURL{%
\tempurl}
\newblock
\shownote{(Retrieved February 2018).}


\bibitem[\protect\citeauthoryear{Gordon, Henzinger, Nori, and Rajamani}{Gordon
  et~al\mbox{.}}{2014}]%
        {gordon_et_al_2014}
\bibfield{author}{\bibinfo{person}{Andrew~D. Gordon},
  \bibinfo{person}{Thomas~A. Henzinger}, \bibinfo{person}{Aditya~V. Nori},
  {and} \bibinfo{person}{Sriram~K. Rajamani}.} \bibinfo{year}{2014}\natexlab{}.
\newblock \showarticletitle{Probabilistic Programming}. In
  \bibinfo{booktitle}{\emph{ICSE track on Future of Software Engineering
  (FOSE)}}. \bibinfo{pages}{167--181}.
\newblock
\urldef\tempurl%
\url{https://doi.org/10.1145/2593882.2593900}
\showURL{%
\tempurl}


\bibitem[\protect\citeauthoryear{Gorinova, Gordon, and Sutton}{Gorinova
  et~al\mbox{.}}{2018}]%
        {gorinova_gordon_sutton_2018}
\bibfield{author}{\bibinfo{person}{Maria~I. Gorinova},
  \bibinfo{person}{Andrew~D. Gordon}, {and} \bibinfo{person}{Charles Sutton}.}
  \bibinfo{year}{2018}\natexlab{}.
\newblock \showarticletitle{{SlicStan}: Improving Probabilistic Programming
  using Information Flow Analysis}. In \bibinfo{booktitle}{\emph{Workshop on
  Probabilistic Programming Languages, Semantics, and Systems (PPS)}}.
\newblock
\urldef\tempurl%
\url{https://pps2018.soic.indiana.edu/files/2017/12/SlicStanPPS.pdf}
\showURL{%
\tempurl}


\bibitem[\protect\citeauthoryear{Gupta, Agrawal, Gopalakrishnan, and
  Narayanan}{Gupta et~al\mbox{.}}{2015}]%
        {gupta_et_al_2015}
\bibfield{author}{\bibinfo{person}{Suyog Gupta}, \bibinfo{person}{Ankur
  Agrawal}, \bibinfo{person}{Kailash Gopalakrishnan}, {and}
  \bibinfo{person}{Pritish Narayanan}.} \bibinfo{year}{2015}\natexlab{}.
\newblock \showarticletitle{Deep learning with limited numerical precision}. In
  \bibinfo{booktitle}{\emph{International Conference on Machine Learning
  (ICML)}}. \bibinfo{pages}{1737--1746}.
\newblock
\urldef\tempurl%
\url{http://proceedings.mlr.press/v37/gupta15.pdf}
\showURL{%
\tempurl}


\bibitem[\protect\citeauthoryear{Hudak}{Hudak}{1998}]%
        {hudak_1998}
\bibfield{author}{\bibinfo{person}{Paul Hudak}.}
  \bibinfo{year}{1998}\natexlab{}.
\newblock \showarticletitle{Modular domain specific languages and tools}. In
  \bibinfo{booktitle}{\emph{International Conference on Software Reuse
  (ICSR)}}. \bibinfo{pages}{134--142}.
\newblock
\urldef\tempurl%
\url{https://doi.org/10.1109/ICSR.1998.685738}
\showURL{%
\tempurl}


\bibitem[\protect\citeauthoryear{Kingma and Welling}{Kingma and
  Welling}{2013}]%
        {kingma_welling_2013}
\bibfield{author}{\bibinfo{person}{Diederik~P. Kingma} {and}
  \bibinfo{person}{Max Welling}.} \bibinfo{year}{2013}\natexlab{}.
\newblock \bibinfo{title}{Auto-encoding variational {Bayes}}.
\newblock   (\bibinfo{year}{2013}).
\newblock
\urldef\tempurl%
\url{https://arxiv.org/abs/1312.6114}
\showURL{%
\tempurl}


\bibitem[\protect\citeauthoryear{Krizhevsky, Sutskever, and Hinton}{Krizhevsky
  et~al\mbox{.}}{2012}]%
        {krizhevsky_sutskever_hinton_2012}
\bibfield{author}{\bibinfo{person}{Alex Krizhevsky}, \bibinfo{person}{Ilya
  Sutskever}, {and} \bibinfo{person}{Geoffrey~E. Hinton}.}
  \bibinfo{year}{2012}\natexlab{}.
\newblock \showarticletitle{ImageNet Classification with Deep Convolutional
  Networks}. In \bibinfo{booktitle}{\emph{Advances in Neural Information
  Processing Systems (NIPS)}}.
\newblock
\urldef\tempurl%
\url{http://papers.nips.cc/paper/4824-imagenet-classification-with-deep-convolutional-neural-networks}
\showURL{%
\tempurl}


\bibitem[\protect\citeauthoryear{Lake, Salakhutdinov, and Tenenbaum}{Lake
  et~al\mbox{.}}{2015}]%
        {lake_salakhutdinov_tenenbaum_2015}
\bibfield{author}{\bibinfo{person}{Brenden~M. Lake}, \bibinfo{person}{Ruslan
  Salakhutdinov}, {and} \bibinfo{person}{Joshua~B. Tenenbaum}.}
  \bibinfo{year}{2015}\natexlab{}.
\newblock \showarticletitle{Human-level concept learning through probabilistic
  program induction}.
\newblock \bibinfo{journal}{\emph{Science}}  \bibinfo{volume}{350}
  (\bibinfo{date}{Dec.} \bibinfo{year}{2015}), \bibinfo{pages}{1332--1338}.
\newblock
Issue 6266.
\urldef\tempurl%
\url{http://science.sciencemag.org/content/350/6266/1332}
\showURL{%
\tempurl}


\bibitem[\protect\citeauthoryear{LeCun, Bengio, and Hinton}{LeCun
  et~al\mbox{.}}{2015}]%
        {lecun_bengio_hinton_2015}
\bibfield{author}{\bibinfo{person}{Yann LeCun}, \bibinfo{person}{Yoshua
  Bengio}, {and} \bibinfo{person}{Geoffrey Hinton}.}
  \bibinfo{year}{2015}\natexlab{}.
\newblock \showarticletitle{Deep Learning}.
\newblock \bibinfo{journal}{\emph{Nature}} \bibinfo{volume}{521},
  \bibinfo{number}{7553} (\bibinfo{date}{May} \bibinfo{year}{2015}),
  \bibinfo{pages}{436--444}.
\newblock
\urldef\tempurl%
\url{https://www.nature.com/articles/nature14539}
\showURL{%
\tempurl}


\bibitem[\protect\citeauthoryear{LeCun, Cortes, and Burges}{LeCun
  et~al\mbox{.}}{1998}]%
        {lecun_cortes_burges_1998}
\bibfield{author}{\bibinfo{person}{Yann LeCun}, \bibinfo{person}{Corinna
  Cortes}, {and} \bibinfo{person}{Christopher~J.C. Burges}.}
  \bibinfo{year}{1998}\natexlab{}.
\newblock \bibinfo{title}{The {MNIST} Database of Handwritten Digits}.
\newblock   (\bibinfo{year}{1998}).
\newblock
\urldef\tempurl%
\url{http://yann.lecun.com/exdb/mnist/}
\showURL{%
\tempurl}
\newblock
\shownote{(Retrieved February 2018).}


\bibitem[\protect\citeauthoryear{Neubig, Dyer, Goldberg, Matthews, Ammar,
  Anastasopoulos, Ballesteros, Chiang, Clothiaux, Cohn, Duh, Faruqui, Gan,
  Garrette, Ji, Kong, Kuncoro, Kumar, Malaviya, Michel, Oda, Richardson,
  Saphra, Swayamdipta, and Yin}{Neubig et~al\mbox{.}}{2017}]%
        {neubig_et_al_2017}
\bibfield{author}{\bibinfo{person}{Graham Neubig}, \bibinfo{person}{Chris
  Dyer}, \bibinfo{person}{Yoav Goldberg}, \bibinfo{person}{Austin Matthews},
  \bibinfo{person}{Waleed Ammar}, \bibinfo{person}{Antonios Anastasopoulos},
  \bibinfo{person}{Miguel Ballesteros}, \bibinfo{person}{David Chiang},
  \bibinfo{person}{Daniel Clothiaux}, \bibinfo{person}{Trevor Cohn},
  \bibinfo{person}{Kevin Duh}, \bibinfo{person}{Manaal Faruqui},
  \bibinfo{person}{Cynthia Gan}, \bibinfo{person}{Dan Garrette},
  \bibinfo{person}{Yangfeng Ji}, \bibinfo{person}{Lingpeng Kong},
  \bibinfo{person}{Adhiguna Kuncoro}, \bibinfo{person}{Gaurav Kumar},
  \bibinfo{person}{Chaitanya Malaviya}, \bibinfo{person}{Paul Michel},
  \bibinfo{person}{Yusuke Oda}, \bibinfo{person}{Matthew Richardson},
  \bibinfo{person}{Naomi Saphra}, \bibinfo{person}{Swabha Swayamdipta}, {and}
  \bibinfo{person}{Pengcheng Yin}.} \bibinfo{year}{2017}\natexlab{}.
\newblock \bibinfo{title}{{DyNet}: The Dynamic Neural Network Toolkit}.
\newblock   (\bibinfo{year}{2017}).
\newblock


\bibitem[\protect\citeauthoryear{Niu, Recht, R{\'e}, and Wright}{Niu
  et~al\mbox{.}}{2011}]%
        {niu_et_al_2011}
\bibfield{author}{\bibinfo{person}{Feng Niu}, \bibinfo{person}{Benjamin Recht},
  \bibinfo{person}{Christopher R{\'e}}, {and} \bibinfo{person}{Stephen~J.
  Wright}.} \bibinfo{year}{2011}\natexlab{}.
\newblock \showarticletitle{Hogwild: A Lock-Free Approach to Parallelizing
  Stochastic Gradient Descent}. In \bibinfo{booktitle}{\emph{Conference on
  Neural Information Processing Systems (NIPS)}}. \bibinfo{pages}{693--701}.
\newblock
\urldef\tempurl%
\url{http://papers.nips.cc/paper/4390-hogwild-a-lock-free-approach-to-parallelizing-stochastic-gradient-descent}
\showURL{%
\tempurl}


\bibitem[\protect\citeauthoryear{Rezende, Mohamed, Danihelka, Gregor, and
  Wierstra}{Rezende et~al\mbox{.}}{2016}]%
        {rezende_et_al_2016}
\bibfield{author}{\bibinfo{person}{Danilo~J. Rezende}, \bibinfo{person}{Shakir
  Mohamed}, \bibinfo{person}{Ivo Danihelka}, \bibinfo{person}{Karol Gregor},
  {and} \bibinfo{person}{Daan Wierstra}.} \bibinfo{year}{2016}\natexlab{}.
\newblock \showarticletitle{One-shot Generalization in Deep Generative Models}.
  In \bibinfo{booktitle}{\emph{International Conference on Machine Learning
  (ICML)}}. \bibinfo{pages}{1521--1529}.
\newblock
\urldef\tempurl%
\url{http://proceedings.mlr.press/v48/rezende16.html}
\showURL{%
\tempurl}


\bibitem[\protect\citeauthoryear{Rezende, Mohamed, and Wierstra}{Rezende
  et~al\mbox{.}}{2014}]%
        {rezende_mohamed_wiestra_2014}
\bibfield{author}{\bibinfo{person}{Danilo~J. Rezende}, \bibinfo{person}{Shakir
  Mohamed}, {and} \bibinfo{person}{Daan Wierstra}.}
  \bibinfo{year}{2014}\natexlab{}.
\newblock \showarticletitle{Stochastic Backpropagation and Approximate
  Inference in Deep Generative Models}. In
  \bibinfo{booktitle}{\emph{International Conference on Machine Learning
  (ICML)}}. \bibinfo{pages}{1278--1286}.
\newblock
\urldef\tempurl%
\url{http://proceedings.mlr.press/v32/rezende14.html}
\showURL{%
\tempurl}


\bibitem[\protect\citeauthoryear{Rumelhart, Hinton, and Williams}{Rumelhart
  et~al\mbox{.}}{1986}]%
        {rumelhart_hinton_williams_1986}
\bibfield{author}{\bibinfo{person}{David~E. Rumelhart},
  \bibinfo{person}{Geoffrey~E. Hinton}, {and} \bibinfo{person}{Ronald~J.
  Williams}.} \bibinfo{year}{1986}\natexlab{}.
\newblock \showarticletitle{Learning representations by back-propagating
  errors}.
\newblock \bibinfo{journal}{\emph{Nature}} \bibinfo{number}{323}
  (\bibinfo{date}{Oct.} \bibinfo{year}{1986}), \bibinfo{pages}{533--536}.
\newblock
\urldef\tempurl%
\url{https://doi.org/doi:10.1038/323533a0}
\showURL{%
\tempurl}


\bibitem[\protect\citeauthoryear{Salvatier, Wiecki, and Fonnesbeck}{Salvatier
  et~al\mbox{.}}{2016}]%
        {pymc3}
\bibfield{author}{\bibinfo{person}{John Salvatier}, \bibinfo{person}{Thomas~V.
  Wiecki}, {and} \bibinfo{person}{Christopher Fonnesbeck}.}
  \bibinfo{year}{2016}\natexlab{}.
\newblock \showarticletitle{Probabilistic programming in Python using PyMC3}.
\newblock \bibinfo{journal}{\emph{PeerJ Computer Science}}  \bibinfo{volume}{2}
  (\bibinfo{date}{April} \bibinfo{year}{2016}), \bibinfo{pages}{e55}.
\newblock
\showISSN{2376-5992}
\urldef\tempurl%
\url{https://doi.org/10.7717/peerj-cs.55}
\showDOI{\tempurl}


\bibitem[\protect\citeauthoryear{Siddharth, Paige, van~de Meent, Desmaison,
  Goodman, Kohli, Wood, and Torr}{Siddharth et~al\mbox{.}}{2017}]%
        {siddharth_et_al_2017}
\bibfield{author}{\bibinfo{person}{N. Siddharth}, \bibinfo{person}{Brooks
  Paige}, \bibinfo{person}{Jan-Willem van~de Meent}, \bibinfo{person}{Alban
  Desmaison}, \bibinfo{person}{Noah~D. Goodman}, \bibinfo{person}{Pushmeet
  Kohli}, \bibinfo{person}{Frank Wood}, {and} \bibinfo{person}{Philip Torr}.}
  \bibinfo{year}{2017}\natexlab{}.
\newblock \showarticletitle{Learning Disentangled Representations with
  Semi-Supervised Deep Generative Models}. In
  \bibinfo{booktitle}{\emph{Advances in Neural Information Processing Systems
  (NIPS)}}. \bibinfo{pages}{5927--5937}.
\newblock
\urldef\tempurl%
\url{http://papers.nips.cc/paper/7174-learning-disentangled-representations-with-semi-supervised-deep\\-generative-models.pdf}
\showURL{%
\tempurl}


\bibitem[\protect\citeauthoryear{Tran, Hoffman, Saurous, Brevdo, Murphy, and
  Blei}{Tran et~al\mbox{.}}{2017}]%
        {tran_et_al_2017}
\bibfield{author}{\bibinfo{person}{Dustin Tran}, \bibinfo{person}{Matthew~D.
  Hoffman}, \bibinfo{person}{Rif~A. Saurous}, \bibinfo{person}{Eugene Brevdo},
  \bibinfo{person}{Kevin Murphy}, {and} \bibinfo{person}{David~M. Blei}.}
  \bibinfo{year}{2017}\natexlab{}.
\newblock \showarticletitle{Deep Probabilistic Programming}. In
  \bibinfo{booktitle}{\emph{International Conference on Learning
  Representations (ICLR)}}.
\newblock
\urldef\tempurl%
\url{https://arxiv.org/abs/1701.03757}
\showURL{%
\tempurl}


\bibitem[\protect\citeauthoryear{Uber}{Uber}{2017}]%
        {pyro}
\bibfield{author}{\bibinfo{person}{Uber}.} \bibinfo{year}{2017}\natexlab{}.
\newblock \bibinfo{title}{Pyro}.
\newblock   (\bibinfo{year}{2017}).
\newblock
\urldef\tempurl%
\url{http://pyro.ai/}
\showURL{%
\tempurl}
\newblock
\shownote{(Retrieved February 2018).}


\bibitem[\protect\citeauthoryear{Wu, Schuster, Chen, Le, Norouzi, Macherey,
  Krikun, Cao, Gao, Macherey, Klingner, Shah, Johnson, Liu, Kaiser, Gouws,
  Kato, Kudo, Kazawa, Stevens, Kurian, Patil, Wang, Young, Smith, Riesa,
  Rudnick, Vinyals, Corrado, Hughes, and Dean}{Wu et~al\mbox{.}}{2016}]%
        {wu_et_al_2016}
\bibfield{author}{\bibinfo{person}{Yonghui Wu}, \bibinfo{person}{Mike
  Schuster}, \bibinfo{person}{Zhifeng Chen}, \bibinfo{person}{Quoc~V. Le},
  \bibinfo{person}{Mohammad Norouzi}, \bibinfo{person}{Wolfgang Macherey},
  \bibinfo{person}{Maxim Krikun}, \bibinfo{person}{Yuan Cao},
  \bibinfo{person}{Qin Gao}, \bibinfo{person}{Klaus Macherey},
  \bibinfo{person}{Jeff Klingner}, \bibinfo{person}{Apurva Shah},
  \bibinfo{person}{Melvin Johnson}, \bibinfo{person}{Xiaobing Liu},
  \bibinfo{person}{Lukasz Kaiser}, \bibinfo{person}{Stephan Gouws},
  \bibinfo{person}{Yoshikiyo Kato}, \bibinfo{person}{Taku Kudo},
  \bibinfo{person}{Hideto Kazawa}, \bibinfo{person}{Keith Stevens},
  \bibinfo{person}{George Kurian}, \bibinfo{person}{Nishant Patil},
  \bibinfo{person}{Wei Wang}, \bibinfo{person}{Cliff Young},
  \bibinfo{person}{Jason Smith}, \bibinfo{person}{Jason Riesa},
  \bibinfo{person}{Alex Rudnick}, \bibinfo{person}{Oriol Vinyals},
  \bibinfo{person}{Greg Corrado}, \bibinfo{person}{Macduff Hughes}, {and}
  \bibinfo{person}{Jeffrey Dean}.} \bibinfo{year}{2016}\natexlab{}.
\newblock \bibinfo{title}{Google's neural machine translation system: Bridging
  the gap between human and machine translation}.
\newblock   (\bibinfo{year}{2016}).
\newblock
\urldef\tempurl%
\url{https://arxiv.org/abs/1609.08144}
\showURL{%
\tempurl}


\end{thebibliography}

\end{document}